\pdfoutput=1

\documentclass[11pt]{article}

\usepackage[]{acl}

\usepackage{times}
\usepackage{latexsym}

\usepackage[T1]{fontenc}

\usepackage[utf8]{inputenc}

\usepackage{microtype}
\usepackage{pifont}
\usepackage[font=small]{caption}
\usepackage[]{subcaption}
\usepackage{verbatim}
\usepackage{framed}
\usepackage{color, colortbl}
\usepackage[colorinlistoftodos]{todonotes}
\usepackage{amssymb}
\usepackage{amsmath}
\usepackage{breqn}
\usepackage{capt-of}
\usepackage{multirow}
\usepackage{algorithmicx}
\usepackage{algorithm}
\usepackage{algcompatible}
\usepackage[normalem]{ulem}
\usepackage{paralist}
\usepackage{enumitem}
\usepackage{balance}
\usepackage{tabularx}
\usepackage[all]{nowidow}
\usepackage{bm}
\usepackage{diagbox}
\usepackage{hyperref}
\usepackage{breqn}

\definecolor{high}{HTML}{FFCE8E}

\title{\textsc{CTM} - A Model for Large-Scale Multi-View Tweet Topic Classification}

\author{Vivek Kulkarni \\
    Twitter Cortex\\ 
  {\tt vkulkarni@twitter.com} \\ \And
  Kenny Leung \\
  Twitter Cortex \\ 
  {\tt kennyleung@twitter.com} \\ \AND
  Aria Haghighi\\
  Twitter Cortex \\ 
  {\tt ahaghighi@twitter.com} \\
}

\newif\ifreviewer

\reviewerfalse

\ifreviewer 
\newcommand{\meta}[1]{\textcolor{brown}{}}
\newcommand{\vivek}[1]{\textcolor{cyan}{$_{vivek}$[#1]}}
\newcommand{\aria}[1]{\textcolor{red}{$_{aria}$[#1]}}

\else
\newcommand{\meta}[1]{\textcolor{brown}{}}
\newcommand{\vivek}[1]{\textcolor{cyan}{}}
\newcommand{\aria}[1]{\textcolor{red}{}}

\fi

\allowdisplaybreaks
\DeclareCaptionFont{10pt}{\fontsize{10pt}{12pt}\selectfont}
\newcommand{\ourmodel}{\textsc{CTM}}

\date{}

\begin{document}
\maketitle
\begin{abstract}
Automatically associating social media posts with topics is an important prerequisite for effective search and recommendation on many social media platforms. However, topic classification of such posts is quite challenging because of (a) a large topic space (b) short text with weak topical cues and (c) multiple topic associations per post. In contrast to most prior work which only focuses on post classification into a small number of topics ($10$-$20$), we consider the task of large-scale topic classification in the context of Twitter where the topic space is $10$ times larger with potentially multiple topic associations per Tweet. We address the challenges above by proposing a novel neural model, \ourmodel\ that (a) supports a large topic space of $300$ topics and (b) takes a holistic approach to tweet content modeling -- leveraging multi-modal content, author context, and deeper semantic cues in the Tweet.  Our method offers an effective way to classify Tweets into topics at scale by yielding superior performance to other approaches (a relative lift of $\mathbf{20}\%$ in median average precision score) and has been successfully deployed in production at Twitter. 

\end{abstract}

\section{Introduction}
On many social media platforms like Twitter, users find posts that they are interested in through two mechanisms: (a) search and (b) recommendation. Both mechanisms typically use the topics associated with posts to identify potential candidates that are displayed to the user. Therefore, automatically associating a post with topics is important for effective search and recommendation. Furthermore, due to the diverse nature of social media content, for such topic association to be useful in practice, it is important to (a) support classification into a large number of topics (potentially hundreds or thousands of topics) and (b) allow for a post to have multiple topics or no topic at all. 

Traditionally, there has been a long line of work on classifying documents (like news articles, movie reviews etc.) into topics ~\cite{borko1963automatic, balabanovic1995learning,joachims1998text,tsutsumi2007movie, yang2014large,adhikari2019docbert}. Additionally, there have been attempts to leverage known label hierarchy to perform hierarchical classification of documents. Most of these approaches learn a model per node of the hierarchy with potentially some form of hierarchy-based regularization in-order to assign labels to a document at each level in the label taxonomy~\cite{koller1997hierarchically, gopal2013recursive,rojas2020efficient}. With the rise of social media platforms, researchers noted that classification of social media content poses several unique challenges ~\cite{chang2015got}. First, such posts can be very short and noisy with very weak cues provided by the linguistic context ~\cite{baldwin2013noisy}. Second, content may be multi-modal with associated images, videos, and hyperlinks. Approaches for classifying documents tend to ignore this multi-modal nature ~\cite{chang2015got}. Several works do explore classification of social media posts (like Tweets) ~\cite{lee2011twitter, genc2011discovering,tao2012makes, stavrianou2014nlp, selvaperumal2014short, cordobes2014graph, kataria2015supervised, chang2015got, li2016Tweet, li2016Tweetsift,li2016hashtag,ive2018hierarchical, kang2019deep, gonzalez2021twilbert}. However, all of these works suffer from one or more limitations: (a) support only a few topics (an order of $10$ topics) (b) model only the text, ignore multi-modal content, deeper semantic-cues and (c) do not support multiple labels per post.

In this paper, we address all of the above limitations in the context of Tweet classification. We propose \ourmodel\ (Concept Topic Model), a Tweet topic classification model that (a) supports classification into $300$ topics ($10$ times larger than prior work) (b) incorporates rich content like media, hyperlinks, author features, entity features thus moving beyond shallow Tweet text features and (c) supports multiple topics to be associated per Tweet.  Our method offers an effective way to classify Tweets into topics at scale and is superior in performance to other approaches yielding a significant relative lift of $\mathbf{20}\%$ in median average precision score. \ourmodel\ has been successfully deployed at Twitter where online A/B experiments have also shown increased engagement and improved customer experience. 

\section{Related Work}
 Early works on Tweet classification used bag-of-words features constructed from Tweet text and classifiers like Rocchio classifiers, logistic regression, and support-vector machines ~\cite{lee2011twitter, genc2011discovering,tao2012makes,stavrianou2014nlp, selvaperumal2014short}.  Follow-up work investigated using increasingly rich features for topic classification including graph-based features of term-co-occurrence graphs, hyperlink information, and distributed representations derived from deep learning models \cite{cordobes2014graph,kataria2015supervised,  li2016using, li2016Tweet,li2016Tweetsift,li2016hashtag, ive2018hierarchical,kang2019deep, gonzalez2021twilbert}. 
 
However, one notes at-least one of the following limitations in all of the above works: (a) focus on a very small number of topics ($5-20$) (b) do not support multiple topic labels per Tweet (c) do not consider or discuss how to model content beyond the raw Tweet text (d) do not capture label constraints. A sole exception to some of the above limitations is the work of \citet{yang2014large} which performs large-scale Tweet topic classification focusing on $300$ topic labels in a real-time setting using only n-gram based features derived from the Tweet text, but ignores other cues. We revisit their large-scale setting after a decade and propose a vastly improved model for large-scale Tweet topic classification modeling Tweets holistically.

\section{Data}
Similar to ~\citet{yang2014large}, we consider a set of $300$ popular Twitter Topics \footnote{We focus on only English Tweets. See the Appendix for the full list of topics considered.}.  While ~\citet{yang2014large} construct  data by only using weak labels obtained from a rule-based system using keyword matches, we employ both high precision human-labeled annotations and weakly-labeled data from a rule-based system using keyword matches \footnote{See the Appendix for a brief description of this rule-based system for yielding weak labels.} to construct the following datasets:
\begin{itemize}[nosep]
\item \textbf{Human Labeled Data (HCOMP Dataset)}:
 We closely follow the procedure outlined by \citet{yang2014large} which first samples Tweets based on topic priors to obtain Tweets that are weakly relevant to a topic, and then seeks label confirmation from trained human annotators. Specifically, we consider Tweets originating from users that are known to tweet mostly about a given topic (for example: Tweets authored by CNN are almost certainly about the ``News'' topic). We collect $100$K such Tweets  with at-least $200$ Tweets per topic. We then sought label confirmation from trained human annotators with each Tweet-topic pair being independently rated by $3$ annotators and use a majority vote to determine the final labels (see Appendix for details). Finally, we create training, validation, and test splits of this dataset disjoint at both the Tweet and the user level.\footnote{We do this because as we will see later, we use author level features in our model.}

\item \textbf{Weakly Labeled Data (WLD Dataset)}:
 We also construct a large-scale data-set of weakly labeled Tweets (\textbf{WLD} dataset) for task-specific pre-training (see Section \ref{sec:models}). Specifically, we use the rule-based system to obtain a random sample of $250$ million weakly labeled Tweets that is disjoint from the \textbf{HCOMP} dataset both in terms of time-span and Tweets. 

\item \textbf{Chatter Data (CHT Dataset)}:
To ensure that our model does not incorrectly assign topics to what is termed ``Twitter chatter'' -- Tweets that are largely about daily status updates, greetings and clearly non-topical content, we closely follow ~\citet{yang2014large} and construct a dataset of weakly labeled non-topical Tweets by sampling Tweets that trigger none of the topical rules in the rule-based system. We verify that a random sample ($N=150$) (denoted by \textbf{CHT-test}) are indeed non-topical through independent human annotators which we set-aside for model evaluation. The remaining portion also user-disjoint ($N=100000$) is used as training data.
\end{itemize}

\section{Models and Methods}
\label{sec:models}
 \begin{figure}[tb!]
\centering
    \includegraphics[width=0.35\textwidth]{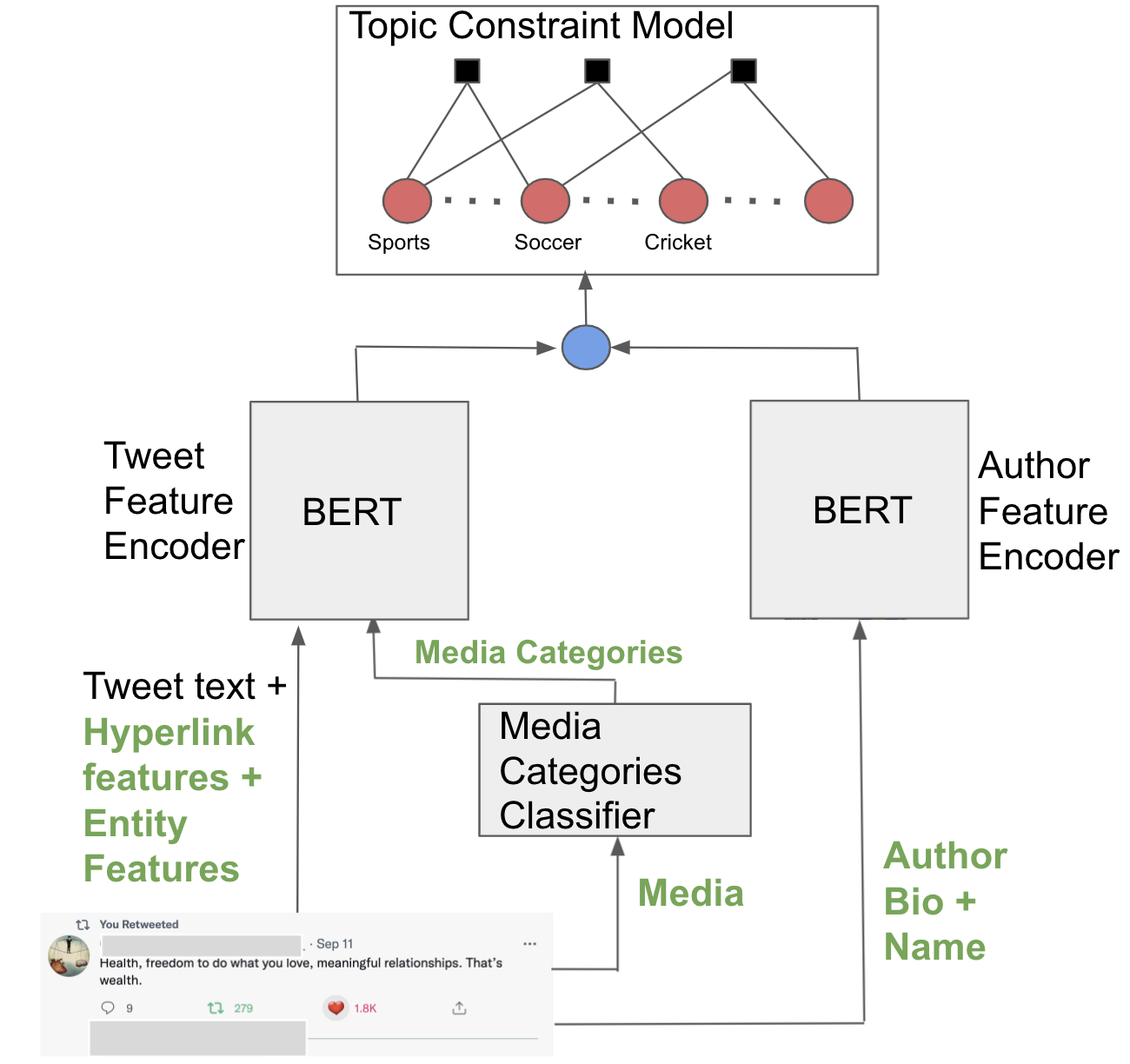}
	\caption{\small{\textbf{Overview of our \ourmodel\ model for large-scale topic classification of Tweets}. Our model consists of $3$ components: (a) a Tweet feature encoder encoding Tweet features (b) an Author feature encoder encoding author features thus capturing author-topic affinity and (c) a constraint model that encourages the topic scores to respect prior constraints.}}
	\label{fig:crown_jewel}
\end{figure}

\paragraph{Problem Formulation.} We formulate our problem as one of standard multi-label classification. Formally, let $\mathcal{S}$ denote the given set of topics. Given $\bm{X}$, a set of Tweet features and a set of topics $\bm{L} \in 2^{\mathcal{S}}$, we seek to model $\Pr(\bm{L}|\bm{X})$. We encode the topic labels $\bm{L}$ as a binary vector $\bm{Y}$ of length $|\mathcal{S}|$ using multi-hot encoding. We consider a simple approach to multi-label classification\footnote{We largely consider a flat classification setting given the absence of well-defined, comprehensive and highly agreed-upon topic taxonomy for Twitter topics, and also because this formulation is better aligned with model deployment constraints.} -- a neural architecture parameterized by $\bm{\Theta}$ that outputs a vector $\bm{\hat{Y}}$ of length $|\mathcal{S}|$ where $\bm{\hat{Y}_{i}}\in [0,1]$ is the probability of the Tweet belonging to topic $i$. 

\paragraph{Model Overview.} \ourmodel\ has three components:
\begin{itemize}[nosep]
    \item \textbf{Tweet Feature Encoder}: This component encodes features of the Tweet holistically. Specifically, it encodes the Tweet text, hyperlink features, named entity mention features, as well as features of associated media. This encoder outputs a vector of topic logits (one for each topic) based on these input features which we denote by $\bm{\hat{Y}}^{t}$. 
    
    \item \textbf{Author Feature Encoder}: This component encodes author features like the author name and biography which may be indicative of the author's affinity to certain topics. This encoder outputs a vector of topic logits (one for each topic) based on these input features which we denote by $\bm{\hat{Y}}^{a}$.  $\bm{\hat{Y}}^{a}$ is combined with $\bm{\hat{Y}}^{t}$ via a element-wise addition to yield the combined topic logits -- $\bm{\hat{Y}}^{c}$ which can be converted to  probability scores using a sigmoid transformation. 

    \item \textbf{Topic Constraint Model}: The topic constraint model encourages the predictions to reflect known constraints among the topic labels. For example,  Tweets about ``Soccer'' are almost certainly also about ``Sports'' but very unlikely to also be about ``Basketball''. We encode such pre-specified label constraints in the output space via a factor-graph. Performing inference on the factor-graph re-calibrates the raw probabilities given by $\bm{\hat{Y}}^{c}$ to better reflect the output label constraints yielding the final predicted probabilities for each topic  $\bm{\hat{Y}}^{f}$.  
\end{itemize}

\subsection{Tweet Feature Encoder}
The Tweet feature encoder is a standard \textsc{Bert} encoder with a linear classification head where all layers are trainable. Each individual Tweet feature is modeled as follows:
\begin{itemize}[nosep]
    \item \textbf{Tweet Text}: We simply pass the Tweet text as an input string to \textsc{Bert} after standard pre-processing (case-folding, stripping hyperlinks and user mentions). 
    \item \textbf{Hyperlink Features}: For each hyperlink in the Tweet text, we obtain the raw HTML content of the web-page being referenced, and extract the web-page title and the first $100$ characters of the web-page description. These features are simply concatenated with the Tweet text using a pre-defined separator token. 
    \item \textbf{Media}: To incorporate topical cues from any attached media (images, gifs, and videos), we obtain media annotations for the given media. These media annotations are broad categories that summarize the content of the media.  We then simply concatenate all of these media annotations to the current input string using a pre-defined token as a delimiter. The media annotations themselves are predicted by a media-annotations classifier that learns to assign each media to zero or more categories from a set of pre-defined categories. \footnote{See the Appendix for details on the media categories classifier.} 
    \item \textbf{Entity Features}: Noting that mentions of named entities provide strong topical cues, we extract such mentions in the Tweet text using an off-the-shelf Twitter NER model \cite{mishra-haghighi-2021-improved} and link each extracted named entity to their entry in \textsc{WikiData} where available. We use the \textsc{WikiData} descriptions of each linked entity as additional inputs to the Tweet feature encoder. As an example, this enables \ourmodel\ to infer that Tweets which mention ``Steve Waugh'' are likely about ``Cricket''. 
 \end{itemize}
 
\paragraph{Pretraining the Tweet Feature Encoder.} Noting that the weights of the standard \textsc{Bert} encoder are not reflective of the domain of Tweets and may represent a poor initialization point during subsequent finetuning,  we pretrain the \textsc{Bert} encoder on the task of predicting topics using the \textbf{WLD} dataset only using the raw Tweet text as the input feature. As we will show empirically, this large-scale pre-training improves generalization performance by better adapting the model to Twitter data.

\subsection{Author Feature Encoder}
\meta{Author Feature Encoder -- - predict topic distribution in pretraining -- User bio as input -- User name - -- predict logits.}
The author feature encoder is also identical to a standard \textsc{Bert} encoder with a linear classification head, with all layers being trainable. We use the following features of the author (all of which are simply concatenated together as input to \textsc{Bert}): (a) \textbf{Author Biography}: We use the self-reported publicly available author-profile description of the author posting the Tweet. (b) \textbf{Author Name}: We also use the author's display name. We hypothesize that all of these features may be indicative of the topics that the author likely tweets about.  For example, an author name containing the string ``FashionNews'' strongly suggests that Tweets made by that author will likely be about Fashion.

\subsection{Topic Constraint Model} The topic constraint model encodes output label constraints in the topic prediction and captures correlations among topics. We encode such dependencies via a factor graph. Given a vector of topic predictions (probabilities)  $\bm{\hat{Y}}^{c}$, for each topic $T_{i}$, we associate a discrete binary random variable with that topic $v_{i}$, and a corresponding unary factor with potential function $f_i$ such that $f_{i}(0)= 1.0-\bm{\hat{Y_{i}}}^{c}$ and $f_{i}(1)=\bm{\hat{Y_{i}}}^{c}$. For every constraint between a pair of topics $(i, j)$, we construct a binary factor with potential function $\phi_{i,j}(v_{i}, v_{j})$. This potential function encodes the compatibility between prediction scores for topic $i$ and topic $j$. Domain experts can craft their own potential functions to reflect positive or negative compatibility between topic pairs or alternatively even learn these from correlation data. \ourmodel\ considers two types of constraints:
\begin{itemize}[noitemsep]
    \item \textbf{Broader Topic Inclusion}: If a Tweet is about a specific topic $c$, then it is very likely that the Tweet is also about topic $p$ where $p$ subsumes topic $c$. Other cases are a ``don't-care''.  For example, if a Tweet is about ``Basketball'', it is almost certainly about ``Sports''.  We use the following potential matrix\footnote{The potential matrices are not necessarily unique and other equivalent matrices may exist.} for encoding this type of constraint:
\[
\small
    \begin{array}{r|cc}
        \hbox{\diagbox{\textbf{p}}{\textbf{c}}} & \mathbf{0} & \mathbf{1} \\ \hline
        \mathbf{0} & 0.5  & 0.0 \\ 
        \mathbf{1} & 0.5 & 10.0 \\ 
    \end{array} 
\]
    
    \item \textbf{Topic Pair Exclusion}: At-most one among topic $a$ and $b$ can be active at any time.  For example, it is very unlikely to have a Tweet which is about both Cricket and Basketball. We use the following potential matrix for encoding this type of constraint:
\[
\small
    \begin{array}{r|cc}
        \hbox{\diagbox{\textbf{a}}{\textbf{b}}} & \mathbf{0} & \mathbf{1} \\ \hline
        \mathbf{0} & 0.5  &  0.5 \\ 
        \mathbf{1} & 0.5 & 0 \\ 
    \end{array} 
\]
\end{itemize} 

After constructing a factor graph encoding the specified output constraints, we perform belief propagation\footnote{See the Appendix for more details on this procedure.} on the factor graph to obtain the final marginal probabilities $\bm{\hat{Y}}^{f}$ which reflect the encoded output constraints. In our experiments,  we impose the above constraint types on specific topics falling under (and including) the broad topics of Sports, Music, Animation, Science, Animals, Anime \& Manga. 



\section{Experiments}
\subsection{Quantitative Evaluation}
\paragraph{Baselines and Evaluation Setup.} We consider two baselines: (a) A bag-of-words logistic regression (LR) model -- our best-effort attempt to reproduce the decade old setup of \citet{yang2014large} and (b) a standard \textsc{Bert} model using only the Tweet text thus replacing logistic regression in (a) with a current state of the art deep-learning model. We train all models on the training data set using class weighted binary cross entropy loss, and evaluate them on the two held-out test sets:
\begin{itemize}[nosep]
    \item \textbf{HCOMP Test Set}: We evaluate model performance on the held out test split from the \textbf{HCOMP} dataset. We report the median average precision score over all topics. We consider the average precision score, since unlike the F1 score, it summarizes model performance over all operating thresholds.
    \item \textbf{CHT Test Set}: In order to measure the ability of our models to effectively reject assigning topics to ``non-topical'' Tweets (chatter), we evaluate our models on the held-out chatter test set. Here, we report the number of predictions made by the model over a given probability threshold (lower scores are better).
\end{itemize}
We perform a systematic feature ablation study of our proposed \textsc{CTM} model to quantify the effect of feature sets considered. Table \ref{tab:main} shows the results of our evaluation where model suffixes represent different ablation settings. Note that our full model significantly outperforms the logistic regression and \textsc{BERT} baselines (\textbf{Median APS}: $\mathbf{67.0}$ vs $\mathbf{54.8}$) and yields a relative improvement of $\mathbf{20}\%$ thus underscoring the effectiveness of our approach. We also make the following additional observations:

\begin{table*}[htb!]
\centering
\small
\begin{tabular}{llll}
                   & \textbf{Setting}                          & \textbf{Median APS} $\uparrow$  & \textbf{CHT} $\downarrow$  \\ \hline
\textsc{LR}(baseline) \cite{yang2014large}       & Tweet text (trained on only HCOMP)                               & 33.0                & 108          \\
\textsc{BERT}(baseline)     & Tweet text (trained on only HCOMP)                               & 54.5                & 254          \\
\textsc{BERT} (baseline)               & Tweet text (trained on HCOMP + CHT)                   & 54.8                & 135 \\ \hline
\textsc{CTM-A}              & Tweet text + media annotation (trained on HCOMP + CHT)  & 54.4                & 121          \\
\textsc{CTM-B}              & \textsc{CTM-A} + pretraining                       & 56.7                & 107          \\
\textsc{CTM-C}              & \textsc{CTM-B} + Hyperlink features                      & 57.2                & 101          \\
\textsc{CTM-D}              & \textsc{CTM-C} + Author features                     & 63.3                & 75           \\
\textsc{CTM-E}              & \textsc{CTM-D} + Entity Linking features           & 66.5                & 80           \\
\rowcolor{high}
\textsc{CTM-F} (Full model) & \textsc{CTM-E} + Constraint model                  & 67.0                & 90 \\ \hline          
\end{tabular}
\caption{\textbf{Performance of \ourmodel\ on the test sets.} The median APS is the median average precision on the \textbf{HCOMP} test set (\emph{higher is better}, $N=10000$) where as \textbf{CHT} column shows the number of model predictions exceeding a probability score of $0.9$ (noting robustness to other thresholds) on the \textbf{CHT} test set (\emph{lower is better}).  \ourmodel\ significantly outperforms baselines and demonstrates the effectiveness of modeling content beyond the immediate Tweet text.}
\label{tab:main}
\end{table*}

\begin{table*}[htb!]
\small
\centering
\begin{tabular}{lrr}
\textbf{Topic}    & \multicolumn{1}{l}{\textbf{APS (w/o constraint model)}} & \multicolumn{1}{l}{\textbf{APS (with constraint model)}} \\ \hline
Animation & 0.64 & 0.71  \\ 
Animals & 0.88 & 0.91 \\
Anime \& manga & 0.66 & 0.84 \\
Music  & 0.41 & 0.70  \\
Sports & 0.69 & 0.89  \\
Science & 0.44 & 0.63 \\ \hline
\end{tabular}
\caption{\textbf{Performance improvements due to the constraint model.} The constraint model yields significant improvements on broader topics (as large as $20$ points). Performance on narrower topics do not change significantly.}
\label{tab:constraints}
\end{table*}

\begin{table*}[htb!]
\small
\centering
\begin{tabular}{p{10cm}|l|l}
\textbf{Tweet Content} & \textbf{Predicted Label} & \textbf{Helpful feature} \\ \hline 
In times of trouble, regression models come to me, speaking words of wisdom  & Data Science & Tweet text \\
Power hitter joins \#yellowstorm \texttt{att:Attached media of cricket bat and gloves} & Cricket & Media Annotations \\
Cameras in USC vs UT stopped working, so it is a podcast now & American Football & Author Bio \\
Revealed: Australia’s stars set to be pulled from IPL \texttt{URL to fox.sports domain} & Cricket & Hyperlink \\
cody ko and noel miller are just ... & Digital creators & Entity features \\ \hline
\end{tabular}
\caption{A few examples of correct model predictions that illustrate the benefit of different feature sets.  Tweets are paraphrased to protect user privacy.}
\label{tab:correct_cases}
\end{table*}

\begin{table*}[htb!]
\small
\centering
\begin{tabular}{p{10cm}|l|l}
\textbf{Tweet Content} & \textbf{Predicted Label} & \textbf{Error Reason} \\ \hline 
In life, you have not seen your best days, you have not run your best race ... & Running & Metaphor  \\
Cheerleading the mob is not going to save ... & Cheerleading & Metaphor \\ 
I am going to have very large drink tonight not sure if whisky or cyanide & Food &  Sarcasm or Irony \\
I need my **** ate & Food & NSFW sense \\
This is a thread 1/5... & No topic & Conversation thread \\  \hline
\end{tabular}
\caption{A few challenging cases for our model. Tweets are paraphrased to protect user privacy.}
\label{tab:challenging_cases}
\end{table*}

\begin{itemize}[nosep]
    \item \textbf{Including non-topical tweets in training improves performance of rejecting chatter} Note that including non-topical tweets in the training data improved the performance of the BERT baseline on the \textbf{CHT} dataset (from $\mathbf{254}$ to $\mathbf{135}$ where lower is better). 
    \item \textbf{Media features have a focused impact.} Adding media annotations overall does not affect the median average precision score significantly (compare row \textbf{CTM-A}: $\mathbf{54.4}$ to row above: $\mathbf{54.8}$). However, we observe that many tweets in the evaluation may not contain media annotations.   When we restricted our evaluation to only the tweets containing media, we observed a significant improvement where the corresponding average precision scores are $\mathbf{71.0}$ \textbf{vs} $\mathbf{58.4}$ respectively. By further computing per-topic performance improvement due to media annotations, we note that media features significantly boost the performance of Automotive, US national news, Anime, and Movies which indeed tend to be media rich, suggesting their focused impact.
    
    \item \textbf{Large-scale pretraining of feature encoders boosts overall performance.}  We observe that pre-training the encoders on domain (and task) specific data is very effective (row \textbf{\textsc{CTM-B}} vs \textbf{\textsc{CTM-A}:Median APS -- $\mathbf{56.7}$ vs $\mathbf{54.4}$}). 
    \item \textbf{Hyperlink features have a focused impact.} Similar to media features, we observe that hyperlink features have a negligible overall impact (see row \textbf{\textsc{CTM-C}:Median APS -- $\mathbf{57.2}$ vs $\mathbf{56.7}$}). However as with media features, when we restricted our evaluation to only those instances with hyperlinks we indeed observe a significant performance gain where the corresponding scores are $\mathbf{92.67}$ \textbf{vs} $\mathbf{83.4}$. Similar to our analysis of media features, a per-topic improvement analysis reveals that hyperlink features most improve the performance on Travel, Movies, Gaming, and US national news which tend to be hyperlink heavy.  
    \item \textbf{Author features significantly boost overall performance.} Author features yield the most benefit overall (see row \textbf{\textsc{CTM-D}:Median APS -- $\mathbf{63.3}$ vs $\mathbf{57.2}$}) thus reaffirming the importance of user-level modeling in NLP tasks.
    \item \textbf{Entity features also significantly boost overall performance.} Similar to author features, the entity features also significantly improve overall performance (see row \textbf{\textsc{CTM-E}:Median APS -- $\mathbf{66.5}$ vs $\mathbf{63.3}$}). Drilling down, we noted that entity linking features most improve the performance on Rap, American football, K-pop, Entertainment News, and Cricket -- all topics whose Tweets are likely to mention sport players, movie stars, and musicians that are suggestive of the topic. 
    \item \textbf{The constraint model significantly boosts the performance of the relevant topics.} Including the constraint model very slightly improves the median average precision score (\textbf{\textsc{CTM-F}}:\textbf{Median APS} $\mathbf{67.0}$ \textbf{vs} $\mathbf{66.5}$). This is expected because the constraint model only affects topics for which constraints were included. Looking at the performance on this subset of topics, we note a significant increase in the average precision score (by as much as $\mathbf{20}$ points) due to reduction in constraint violations -- especially violations of the broader topic inclusion constraint (see Table \ref{tab:constraints}).\footnote{This slight degradation on CHT is due to error propagation of high confidence false positives which occurs to respect the constraints.} 
\end{itemize} 

\subsection{Qualitative Evaluation}
In addition to evaluating our \ourmodel\ quantitatively, we also inspected the model predictions qualitatively to identify instances which (a) reveal the benefits of holistic tweet modeling and (b) highlight challenging cases.  Table \ref{tab:correct_cases} shows a few instances that illustrate the benefit of holistically modeling Tweet content. Note that in ``Power hitter joins \#yellowstorm'', only the attached media (which displays a cricket apparel) is indicative of the topic. Similarly, our model correctly predicts that ``Revealed: Australia's stars set to be pulled from IPL'' is about ``Cricket'' by leveraging topical cues extracted from the linked website's content. Finally, \ourmodel\ correctly infers that the Tweet referencing ``Cody Ko and Noel Miller'' is about ``Digital Creators'' by leveraging named entity cues. 
Finally, we also noted a few systematic failure modes (see Table \ref{tab:challenging_cases}). In particular, our model does not pick up on (a) metaphorical usage of topical words like ``running'' or ``cheer-leading'' (b) sarcasm and irony (c) NSFW senses of certain topical phrases (d) topical content in conversational threads since this requires modeling conversational context. 

\subsection{Online Evaluation}
Finally, we also evaluated \ourmodel\ online by performing an A/B test comprising of $25$ million users in each bucket. To summarize the results of the A/B test briefly, we observed that \ourmodel\ relatively increased: (a) the size of the topic Tweet inventory online by about $4\%$. This translates to about $600$K additional topical Tweets daily that could be surfaced to users based on their topical interests to improve their user experience. (b) precision by $5\%$ and (c)  user engagement by $5.5\%$. In a nut-shell, our online experiments suggested that \ourmodel\ significantly improves the user experience of the Topics product surface in Twitter and has consequently been deployed in production.   

\section{Conclusion}
We revisited the problem of large scale Tweet topic classification posed by \citet{yang2014large} and proposed a model for classifying Tweets into a large set of $300$ topics with improved performance. In contrast to prior work we take a holistic approach to modeling Tweets and model not only the immediate Tweet text, but also associated media, hyperlinks, author context, entity mentions, and incorporate domain knowledge expressed as topic constraints in a principled manner. Our model showed significantly increased engagement and improved customer experience in several online A/B experiments, and it has been deployed into production at Twitter with millions of active users. Finally, while our model and approach has been restricted to Tweet classification, our proposed methods and observations may benefit other social media platforms seeking to classify content into a large number of topics effectively. 

\section*{Ethical Considerations} This paper and the data used within was reviewed as part of Twitter's standard privacy and legal review processes.  No data has been publicly released in relation to this paper. While there is a possibility that the model could be misused, we do not anticipate any new or increased risks over those already present in established prior work and prior models on topic classification.

\section*{Acknowledgments} We would like to acknowledge the contributions of Inom Mirzaev, Chenguang Yu, Laleh Soltan Ghoraie, Kovas Boguta, Jun Ng, Andy Aitken, Olivia Ifrim, Shubhanshu Mishra, Sneha Mehta, Aman Saini, Sijun He, Wayne Krug, and Ali Mollahosseini in supporting this work. We would also like to thank the anonymous reviewers for their comments and suggestions. 
\bibliography{anthology,custom}
\bibliographystyle{acl_natbib}

\clearpage
\appendix

\section{Appendix}
\label{sec:appendix}

\subsection{Details Regarding Off the Shelf Components Used in \ourmodel}
\subsubsection{Media Annotations Classifier}
The media annotations classifier takes as input an image and classifies the image into one or more of $45$ media categories listed in Table \ref{tab:media_categories}. The classifier is essentially a standard \textsc{MobileNet V2} model \cite{sandler2018mobilenetv2} further fine-tuned on a human-labeled curated dataset of $100$K images from Twitter. The operating threshold of the media classifier is set to achieve a precision of about $90\%$ on each topic.\footnote{ For videos, and GIF's each frame is analyzed by the model with the  prediction scores being aggregated using the max operator.} 
 
\subsubsection{Twitter Named Entity Recognizer}
The Twitter NER model is a standard bi-directional LSTM with a CRF layer and detects mentions of persons, places, organizations, and products in a Tweet. The model has been trained on $100$K human annotated labeled tweets \cite{mishra2020assessing} and has a precision of $85\%$ with a recall of $70\%$ on a held-out test set. We link the extracted mention to a potential WikiData candidate as follows: (a) we first construct a set of potential WikiData entity candidates - the set of all entities whose label or alias has a match with the extracted mention (b) link the mention to the top entity candidate obtained by sorting the candidate set in descending order of page view count as the primary key breaking ties using page rank as the secondary key. We use this approach as an expedient choice noting that more sophisticated entity linking approaches can be used. 

\subsubsection{Rule Based System for Generating Weakly Labeled Examples.}
We employ a rule-based system consisting of tens of thousands of rules based on key-words to generate weakly labeled examples. All rules are manually curated and added by domain experts and data specialists. 

\subsection{Hyper-parameter Tuning} 
As is standard practice, we use the validation set ($N=10000$) to perform hyper-parameter tuning. We explored several hyper-parameter settings for the baseline models namely Logistic Regression and BERT to make baseline comparisons strong and compare \ourmodel\ against only the best performing baseline settings. In particular, we explored training for different epochs $(1-10)$ for the \textsc{BERT} baseline.  For the logistic regression baseline, we also tried various settings for the maximum number of iterations of the optimizer ($100-1000$) as well as various values for the strength of the L$2$ regularizer ($C=[0, 1, 10, 100]$).  

For our proposed model \ourmodel, we did not do any specific hyper-parameter tuning and just trained all models for $5$ epochs using 1 A$100$ GPU. 

\subsection{Details on the Human Labeled Annotation Task}
In this section, we briefly describe the human annotation task used for obtaining topic label confirmation used in the construction of the \textbf{HCOMP} dataset.  Each annotator is shown a Tweet, topic pair and asked to judge whether the topic is relevant to the Tweet or not. The instructions are:
\\

\fbox{\begin{minipage}{21em}
\small
\textbf{Task}: In this task, you will be shown a tweet and a topic and asked whether the tweet is 'relevant' for a topic.
\\

\textbf{Topics}:You will be asked to determine if a tweet is relevant for a given topic. 
A “Topic” is a potential subject of conversation that can be identified with a commonly held definition, where mass interest in the subject is not likely to be temporary, e.g. ‘Comedy’ or 'Knitting'  is a topic as it is non-subjective and has a commonly held definition. Purely social tweets like “are you doing okay?” or personal remarks like “I’m having a bad day” are not topical. A Tweet can be popular without being topical. 
\\

\textbf{Question}: The primary question you will be asked is ``Is this tweet about a topic?'', the possible responses are:
Yes - This tweet is primarily about this topic. 
Somewhat - This tweet is related to this topic, but it is not a primary topic of this tweet.
No - This tweet is unrelated to this topic.
Unsure - I don't understand this tweet.
\\

\textbf{Guidelines}: You will first want to make sure you understand the presented topic. If you are unfamiliar with the topic presented in this question, please click on the topic which will take you to a Google search result page. Feel free to click on a few links (news articles or a Wikipedia page) to familiarize yourself with the topic. 
When elements of the tweet can I use to make a judgment?
It can sometimes be challenging to tell what a tweet is about from tweet text alone. In order to determine what the tweet is about you may need to do the following:
Look at replies of a tweet, which might provide additional context by clicking on the tweet. (NOTE: If you can understand the tweet by relying just on the body or author of the tweet, it is fine to not designate replies as being used to make a judgment.)
Google phrases in the tweet text if you are unfamiliar with a mentioned entity or phrase that will help you understand the tweet.
Look at the image, video, or click on any link (including a hashtag) associated with the Tweet, since it may be commenting on this media. If the media is primarily about the topic, the tweet is as well. 
Look at the tweet author's name, profile, public timeline, or linked website if it helps disambiguate tweet content. (NOTE: Please don't use the author alone in making determination, without some other element of the tweet.)

\end{minipage}
}
\\

Each HIT was judged by $3$ independent highly reliable annotators. Finally, we noted that two-way (majority) agreement rate was $86\%$, unanimous agreement was $66\%$ and the topic precision overall was $70\%$ (with ``somewhat'' ratings being counted towards a precision error). 
\subsection{Data Statement}
Here, we outline other aspects of our data as per recommendations outlined in \cite{bender2018data}.

\textsc{Summary} -- We collect a set of Tweet, topic pairs focusing on only English Tweets which we use for predictive modeling and evaluation. 

\textsc{Curation Rationale} -- The rationale for the setup used in data collection was primarily driven by our task (large scale topic classification) and the need for data to a build a predictive model. The size of the data collected was thus influenced by task, available budget, and time available.

\textsc{Language variety} - The tweets were restricted to English only and are from the time range between September 2020 and May 2021. More fine-grained information is not available. 

\textsc{Speaker demographic} -- We do not have any demographic information of the users in this data.  One would expect the demographic information to be similar to the demographics of Twitter users  around the time of data collection.

\textsc{Annotator demographic} -- Human Annotators are primarily native English speakers. No other information is available.

\textsc{Text Characteristics} -- Tweets are short, informal and have at-most $280$ characters. Tweets are generally meant to be engaged with by other Twitter users.

\begin{table*}[hb!]
\centering
\small
\begin{tabular}{lll}
\hline
App Screenshots             & Entertainment Events         & Pets                           \\
Arts and Crafts             & Food                         & Piercing                       \\
Auto Racing                 & American Football            & Running                        \\
Automotive                  & Gambling                     & Single Person                  \\
Baseball                    & Gaming                       & Skateboarding                  \\
Basketball                  & Golf                         & Skiing                         \\
Beauty, Style and Fashion   & Hockey                       & Smoking                        \\
Boxing                      & Home and Garden              & Pharmaceuticals and Healthcare \\
Captioned Images            & Infographics, Text and Logos & Snowboarding                   \\
Comics, Animation and Anime & Martial Arts                 & Soccer                         \\
Cricket                     & Multiple People              & Swimming                       \\
Crowds and Protests         & Nature and Wildlife          & Tennis                         \\
Currency                    & Weapons                      & Travel                         \\
Cycling                     & Other                        & TV Broadcasts                  \\
Drinks                      & Performance Arts             & Weather and Natural Disasters  \\ \hline
\end{tabular}
\caption{List of 45 media categories that make up the label space of the media classifier.}
\label{tab:media_categories}
\end{table*}

\begin{table*}[t]
\centering
\tiny
\begin{tabular}{llll}
\hline
2D animation               & Country music               & Horses                       & Rock climbing               \\
3D animation               & Cricket                     & Hotels                       & Rodeo                       \\
Accounting                 & Cruise travel               & Houston                      & Roleplaying games           \\
Action and adventure films & Cult classics               & Independent films            & Romance books               \\
Adventure travel           & Curling                     & Indie rock                   & Rowing                      \\
Advertising                & Cybersecurity               & Information security         & Rugby                       \\
Agriculture                & Cycling                     & Interior design              & Running                     \\
Air travel                 & Dance                       & Internet of things           & Sailing                     \\
Alternative rock           & Darts                       & Investing                    & Saxophone                   \\
American football          & Data science                & J-pop                        & Sci-fi and fantasy          \\
Animals                    & Databases                   & Jazz                         & Sci-fi and fantasy films    \\
Animated films             & Dating                      & Jewelry                      & Science                     \\
Animation                  & Digital creators            & Job searching and networking & Science news                \\
Animation software         & Documentary films           & Judo                         & Screenwriting               \\
Anime                      & Dogs                        & K-hip hop                    & Sculpting                   \\
Anime \& manga              & Drama films                 & K-pop                        & Sharks                      \\
Antiques                   & Drawing and illustration    & Kaiju                        & Shoes                       \\
Archaeology                & Drums                       & Knitting                     & Shopping                    \\
Architecture               & EDM                         & Lacrosse                     & Skateboarding               \\
Art                        & Economics                   & Language learning            & Skiing                      \\
Artificial intelligence    & Education                   & Latin pop                    & Skin care                   \\
Arts\& culture             & Electronic music            & MMA                          & Small business              \\
Arts \& culture news        & Entertainment               & Makeup                       & Sneakers                    \\
Arts and crafts            & Entertainment news          & Marine life                  & Snooker                     \\
Astrology                  & Environmentalism            & Marketing                    & Soap operas                 \\
Astronauts                 & Esports                     & Martial arts                 & Soccer                      \\
Athletic apparel           & Europe travel               & Mathematics                  & Soccer stats                \\
Augmented reality          & Everyday style              & Men's boxing                 & Soccer transfers            \\
Australian rules football  & Experimental music          & Men's golf                   & Soft rock                   \\
Auto racing                & Famous quotes               & Men's style                  & Softball                    \\
Automotive                 & Fantasy baseball            & Motorcycle racing            & Space                       \\
Aviation                   & Fantasy basketball          & Motorcycles                  & Sporting goods              \\
Backpacking                & Fantasy football            & Movie news                   & Sports                      \\
Badminton                  & Fantasy sports              & Movies                       & Sports news                 \\
Ballet                     & Fashion                     & Movies \& TV                  & Sports stats                \\
Baseball                   & Fashion and beauty          & Museums                      & Startups                    \\
Basketball                 & Fashion business            & Music                        & Storyboarding               \\
Beauty                     & Fashion magazines           & Music festivals              & Street art                  \\
Biographies and memoirs    & Fashion models              & Music industry               & Streetwear                  \\
Biology                    & Fast food                   & Music news                   & Supernatural                \\
Biotech and biomedical     & Fiction                     & Music production             & Surfing                     \\
Birdwatching               & Fighting games              & Musicals                     & Swimming                    \\
Black Lives Matter         & Figure skating              & Mystery and crime books      & Table tennis                \\
Blues music                & Financial services          & National parks               & Tabletop gaming             \\
Board games                & Fintech                     & Nature                       & Tabletop role-playing games \\
Bollywood dance            & Fishing                     & Nature photography           & Tattoos                     \\
Bollywood films            & Fitness                     & Netball                      & Tech news                   \\
Bollywood music            & Folk music                  & Nonprofits                   & Technology                  \\
Bollywood news             & Food                        & Olympics                     & Television                  \\
Books                      & Food inspiration            & Online education             & Tennis                      \\
Bowling                    & Futurology                  & Open source                  & Theater                     \\
Boxing                     & Game development            & Opera                        & Theme parks                 \\
Brazilian funk             & Gaming                      & Organic                      & Thriller films              \\
Business \& finance         & Gaming news                 & Organic foods                & Track \& field               \\
Business media             & Gardening                   & Outdoor apparel              & Trading card games          \\
Business news              & Genealogy                   & Outdoors                     & Traditional games           \\
Business personalities     & Geography                   & Painting                     & Travel                      \\
C-pop                      & Geology                     & Parenting                    & Travel guides               \\
Careers                    & Golf                        & Pets                         & Travel news                 \\
Cartoons                   & Graduate school             & Philosophy                   & Triathlon                   \\
Cats                       & Grammy Awards               & Photography                  & US national news            \\
Cheerleading               & Graphic design              & Physics                      & Veganism                    \\
Chemistry                  & Guitar                      & Podcasts \& radio             & Vegetarianism               \\
Chess                      & Gymnastics                  & Poker                        & Venture capital             \\
Classic rock               & Hair care                   & Pop                          & Video games                 \\
Classical music            & Halloween films             & Pop Punk                     & Visual arts                 \\
Cloud computing            & Handbags                    & Pop rock                     & Volleyball                  \\
Cloud platforms            & Hard rock                   & Progressive rock             & Watches                     \\
College life               & Health news                 & Psychology                   & Weather                     \\
Combat sports              & Heavy metal                 & Punjabi music                & Web development             \\
Comedy                     & Historical fiction          & Punk                         & Weddings                    \\
Comedy films               & History                     & R\&B and soul                 & Weight training             \\
Comics                     & Hockey                      & Rap                          & Women's boxing              \\
Computer programming       & Home \& family               & Reality TV                   & Women's golf                \\
Concept Art                & Home improvement            & Reggae                       & Women's gymnastics          \\
Construction               & Horoscope                   & Reggaeton                    & World news                  \\
Cooking                    & Horror films                & Road trips                   & Wrestling                   \\
Cosplay                    & Horse racing and equestrian & Rock                         & Yoga  \\ \hline                      
\end{tabular}
\caption{List of topics making up our topic space.}
\label{tab:all_topics}
\end{table*}

\subsection{Details on Belief Propagation}
In this section, we provide more details on the procedure of belief propagation used in the topic constraint model component. In belief propagation, messages are alternately passed between variable nodes and factor nodes (until convergence is achieved or a finite number of iterations is completed). A message is simply a vector $\mu$ where the individual components denote the probability of the random variable taking a specific value $x \in \{0,1\}$. The message from a variable $v$ to neighboring factor $f$ on taking a specific value $x$ is given by the following equation:
\begin{equation}
    \mu_{v\longrightarrow f}(x) \propto \prod_{g \in \mathcal{N}(v)\setminus f} \mu_{g\longrightarrow v}(x)
\end{equation}, where $g$ belongs to the set of factor nodes connected to $v$ excluding $f$.  
Similarly, the message from a factor node $f$ to the variable $v$ on the variable taking a specific value $x$ is given by the following:
\begin{equation}
    \mu_{f\longrightarrow v}(x) \propto \sum_{\mathbf{x}:\mathbf{x}_{v}=x} \phi(\mathbf{x})\prod_{u \in \mathcal{N}(f)\setminus v} \mu_{u\longrightarrow f}(\mathbf{x}_{u})   
\end{equation}, where $u$ belongs to the set of variable nodes connected to $f$ excluding $v$.

Finally, after convergence (or a finite number of iterations), the updated marginal probability of variable $v$ taking on a value $x$ is given by $\Pr(v = x) \propto \prod_{g \in \mathcal{N}(v)} \mu_{g\longrightarrow v}(x)$.

\end{document}